# Building a Question and Answer System for News Domain

Narayana Darapaneni
*Director – AIML*
*Great Learning/Northwestern*
University Illinois, USA
darapaneni@gmail.com

Pooja Chetan
*Student – AIML*
*Great Learning*
Bangalore, India
poojabk582@gmail.com

Anwesh Reddy Paduri
*Data Scientist - AIML*
*Great Learning*
Bangalore, India
anwesh@greatlearning.in

Aravind Gaddala
*Student – AIML*
*Great Learning*
Bangalore, India
aravind.gaddala@gmail.com

Garima Tiwari
*Student – AIML*
*Great Learning*
Bangalore, India
garimatiwari200@gmail.com

Sandipan Basu
*Student – AIML*
*Great Learning*
Bangalore, India
sandipan.basu@gmail.com

Sadwik Parvathaneni
*Mentor– AIML*
*Great Learning*
Bangalore, India
sadwik25@gmail.com

*Abstract*— This project attempts to build a Question-Answering system in the News Domain, where Passages will be News articles, and anyone can ask a Question against it. We have built a span-based model using an Attention mechanism, where the model predicts the answer to a question as to the position of the start and end tokens in a paragraph. For training our model, we have used the Stanford Question and Answer (SQuAD 2.0) dataset[1]. To do well on SQuAD 2.0, systems must not only answer questions when possible but also determine when no answer is supported by the paragraph and abstain from answering. Our model architecture comprises three layers- Embedding Layer, RNN Layer, and the Attention Layer. For the Embedding layer, we used GloVe and the Universal Sentence Encoder. For the RNN Layer, we built variations of the RNN Layer including bi-LSTM and Stacked LSTM and we built an Attention Layer using a Context to Question Attention and also improvised on the innovative Bidirectional Attention Layer. Our best performing model which uses GloVe Embedding combined with Bi-LSTM and Context to Question Attention achieved an F1 Score and EM of 33.095 and 33.094 respectively. We also leveraged transfer learning and built a Transformer based model using BERT. The BERT-based model achieved an F1 Score and EM of 57.513 and 49.769 respectively. We concluded that the BERT model is superior in all aspects of answering various types of questions.

## I. INTRODUCTION

*Machine Reading Comprehension(MRC)* is the ability to process text, understand its meaning, and to integrate with what the reader already knows. Fundamental skills required in efficient machine reading comprehension system are knowing meaning of words, ability to understand meaning of a word from context, ability to follow organization of passage and to identify antecedents and references in it, ability to draw inferences from a passage about its contents, ability to identify the main thought of a passage and ability to answer questions in a passage. Question and Answer (QnA) is one of the popular applications of MRC and can be considered a subset of the MRC. In recent years, with the success of machine learning techniques, especially Deep Neural Networks, to process sequential data such as texts, MRC has become an active area in the field of NLP. The Machine Reading Comprehension system aims to answer (A) given a passage (P) and a question (Q). This project aims to solve this in the News domain, where Passages will be News articles, and anyone can ask a Question against it. Such a system has wide application[4]:

- News feeds be it online or offline needs literacy. This system can become a foundation technology on which a voice based question and answers can be delivered. Huge implications in rural areas and specially in times where print media is unreachable
- Multilingual news question and answering system
- Help researchers who are mining news archives[8]

This project will attempt to build a Question Answering system(QnA System) and then attempt to answer a user's queries related to a user provided news articles. The answers will be extractive in nature - focusing on extracting key information from the context (i.e. a paragraph or a document) which the user asks for.

To understand the current and past research in this domain, we reviewed research articles available in the public domain relating to MRC and QnA systems for various types of problems, the alternative approaches they evaluated, data that they chose, algorithms they selected and evaluation methods they used to measure the outputs. In the recent past with the availability of large corpus of training data, sophistication in algorithms, availability of computing resources, the Deep Learning based approaches are being predominantly used. Since the domain we have chosen is News Q&A, we would be focussing on 'Factoid' questions- questions that can be answered with simple facts expressed in short text answers like a name, location, date. Question answering is very dependent on a good training corpus, for without documents containing the answer, there is little any question answering system can do. This is especially true for Deep Neural Networks. The most cited and popular training data sets for QnA Systems are - SQuAD 2.0, Newsqa, CNN Daily Mail[5] and MS MARCO. For training our model, we have used the Stanford Question and Answer (SQuAD 2.0) dataset. We selected this data set since it is a closed dataset, meaning that the Answer to a Question is always a part of the Context and also a continuous span of Context. So the problem of finding an answer can be simplified as finding the start index and the end index of the context that corresponds to the answers. SQuAD 2.0 combines existing SQuAD data with over 50,000 unanswerable questions written adversarially by crowdworkers to look similar to answerable ones. To do well on SQuAD 2.0, systems must not only answer questions when possible, but also determine when no answer is supported by the paragraph and abstain from answering. Hence, SQuAD 2.0 is a challenging natural language understanding task. We used F1 Score and EM as the evaluation metrics for the predictions. Our model architecture is composed of an Embedding Layer, RNN Layer, and an Attention Layer, followed by a Prediction Layer and an Optimizer for backpropagation.

There has been tremendous progress in the field of NLP after the introduction of the BERT framework by Google [6]. This achieved State-of-the-Art results on 11 individual NLP tasks. BERT has inspired many recent NLP architectures, training approaches, and language models, such as Google's TransformerXL, OpenAI's GPT-2, etc. We felt that the Transformer models using BERT would do a good job in addressing News domain-specific questions. Hence, we also leveraged transfer learning and built a Transformer based model using BERT. Though this is the best model that we have, this required the least amount of effort on our part. For this project we concentrated our effort on building the RNN models with the Attention mechanism, which is explained in detail in this document.

## II. MATERIALS AND METHODS

The machine reading comprehension (MRC) task as defined here involves a question $Q = \{q_0, q_1, ..., q_{m-1}\}$ and a passage $P = \{p_0, p_1, ..., p_{n-1}\}$ and aims to find an answer span $A = \{a_{start}, a_{end}\}$ in P. We assume that the answer exists in the passage P as a contiguous text string. Here, m and n denote the number of tokens in Q and P, respectively. The learning algorithm for reading comprehension is to learn a function $f(Q, P) \rightarrow A$. The training data is a set of the query, passage and answer tuples $< Q, P, A >$. We will start by describing how Q and P are processed, followed by a description of our model architecture.

### A. Data Pre-processing

Build Target: The Answer in our Data is always a continuous span of text in the Context. We have represented the Answer as a tuple of Answer Start Token (AS) and Answer End Token(AE). AS is the position of the first answer token in the Context; and AE is the position of the last answer token in Context. This can be visualized as shown in Figure 1

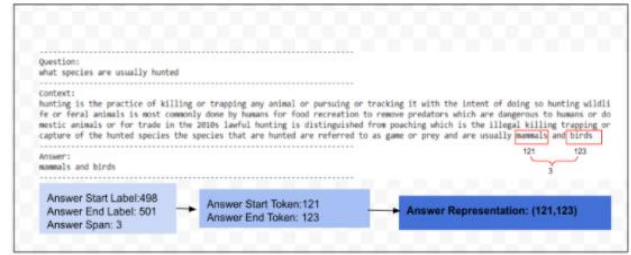

Fig.1. Detecting Answer Span in a Passage

We want to represent the positional index of Answer Start Token and Answer End Token so that we should be able to derive the probability of 'Start Token' and 'End Token' independently. To achieve this we have to represent a Token being as a Start Token in the Context space (with a dimension of Maximum Context length). We have to do this for the End Token also. We concatenate the Start, and End Token encoded vector to represent the Target Variable.

Build Tokenizer: We have built our tokenizer with the following parameters: Number of words= Full Vocabulary size; Tokenization method= Word Tokenizer and Lower Case= True.

Vectorization and Encoding: We performed the following steps as part of Vectorization and Encoding. Converting the context and question texts to sequences. We padded the context and questions to their respective maximum length to convert the sequences to the same length. This vectorization allows code to efficiently perform the matrix operations in batch during the model training. We padded using 0 and applied 'pre-padding' of sequences. Since for our RNN model, we take the final output or hidden state, we would want to ensure that the memory does not get flushed out in the final step.

### B. Model Building

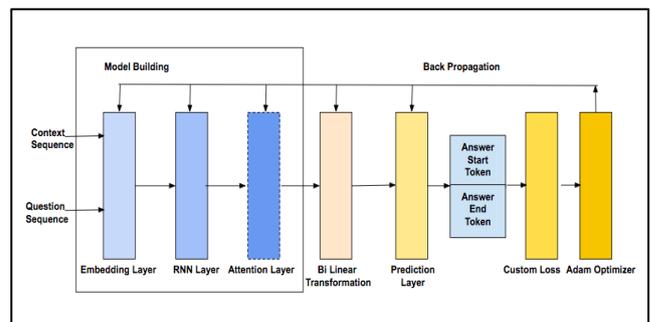

Fig.2. Overview of Model and Architecture

As shown in Figure 2, our model contains different layers to capture different concepts of representations. The detailed description of our model is provided as follows

1. *Input Layer*: The inputs to our model are the padded sequences of Question and Context. The shape of the input sequences: (Vocabulary Size*Max Length of Question or Context Sequence)
2. *Embedding Layer*: We added an Embedding layer for creating word vectors for our Question and Context text sequences. This layer sits between the Input Layer and the RNN Layer. The Inputs to 'Question' Embedding Layer- ( Number of Samples*Max Question or Context Length).The output of the Embedding Layer are vectors of shape: Training Sample Size*Max Length of Question or Context X Embedding Size). We have used two approaches for building the Embedding Layers for the models
3. GloVe Embedding: we used the 300 Dimension Common Crawl for the English language
4. Universal Sentence Encoder: we used the 512 Dimension 'universal-sentence-encoder-qa' which shows strong performance on English language Question and Answer tasks
5. *RNN(LSTM) Layer*: We have built three different model architectures using RNN.
    a. Model 1: Vanilla LSTM Layer: We built a Vanilla LSTM model that has a single layer of LSTM units and an output layer.
    b. Model 2: Bidirectional LSTM Layer: We built a Bidirectional LSTM model to learn the input sequences both forward and backward and concatenate the interpretations to an output layer.
    c. Model 3: Stacked LSTM Layer: We built a Stacked LSTM layer by stacking 3 LSTM layers one on top of another.
6. *Attention Layer*: We have built an Attention Layer, which couples the Question and Context vectors and produces a set of Question aware feature vectors for each word in the Context.We have built two different models with the following two Attention Layers: Context to Question Layer ( C2Q) and Bidirectional Attention Layer (biDAF)- this is a combination of Question to Context Attention and Context to Question Attention. Context to Question Attention Layer ( C2Q) Context-to-question (C2Q) attention signifies which Question words are most relevant to each Context word. We build this layer using the following steps
a. Alignment Weights - We compute the Alignment Weights through a dot product of the hidden states of the Question Vectors and Context Vectors
b. Alignment Matrix - We derive the Alignment Matrix by inputting the alignment weights through a softmax layer. We use a softmax layer to convert the weights into a set of probabilities of the Alignment weights. This helps identify the context words that are most relevant to the context words.
c. Question Attention Vector - We build the Question Vector through a dot product of the alignment matrix with the hidden states of the Question Vectors
d. Context Attention Vector - We build this by concatenating the Question Attention Vector with the Context LSTM output
e. C2Q Attention Vector - We finally build the Attention Vector by inputting the Context Attention Vector through a Dense layer with tanh activation.
7. The output of the Attention layer is the Question aware vector representations of the Context words.
8. Bidirectional Attention Layer (biDAF)- We have taken inputs from this research Paper [3].This original model uses a hierarchical multi-stage architecture to model the representations of the context paragraph at different granularity levels. It includes character-level, word-level, and contextual embeddings, and uses bi-directional attention flow to obtain a query-aware context representation. The intuition for building the biDirectional Attention Layer is that by evaluating both the similarity and relevance of each context word with the question word, the Context is more aware of the Question. We have made changes to the Original model and have used only word-level embeddings. We built a combination of Question to Context Attention(Q2C) and Context to Question(C2Q) Attention Layers. C2Q attention signifies which Question words are most relevant to each Context word, and Q2C attention means which Context words have the closest similarity to one of the Question words and are hence critical for answering the query. First, we model the C2Q attention layer, as described above. Next, we model the Q2C attention layer. The steps for building the Q2C attention are similar to that for the C2Q attention layer.
a. Alignment Weights - We compute the Alignment Weights through a dot product of the hidden states of the Question Vectors and Context Vectors
b. Alignment Matrix- After we compute the Alignment Weights through a dot product of the hidden states of the Question Vectors and Context Vectors, we compute the 'max score' for the Alignment weights before passing the alignment weights through the softmax layer. This operation ensures that this step's output represents a tensor- which is the most important context word for the given question. This is the small difference in the modelling of this attention layer compared to the C2Q attention layer.
c. Context Attention Vector- After we compute the Alignment matrix, we build a Context Attention Vector through a dot product of the Alignment matrix with the Context Vectors' hidden states.
d. Attention Vector- This context vector is then tiled across all context dimensions to build the Q2C Attention Layer.
e. Bidirectional Attention Layer- the last step is to merge these two attention layers(C2Q and Q2C) to build the biDirectional Attention layer.
f. *Bi-linearity Transformation:* We use the Bilinear Transformation to capture the similarity between each Context Token and Question. The input to the bilinear transformation layer is the output from the previous RNN Layer or Attention Layer. The output is a matrix with the probability for a token (from the Context vector) being the Start token and End token[10]. .

9. *Prediction Layer*: The purpose of the Prediction layer is to predict the position of two Tokens(start token and end token) in the Context Vector that together have the maximum probability of being the correct Answer for a given Question. The inputs to the Prediction Layer from the previous biLinear Transformation are the probabilities of every token in the Context being the start and end token for a given question. We performed the following steps to model the Prediction Layer:
a. We compute the joint probability of a token being the start token and token being the end token. We do this by multiplying the probability(start_probab) of a token(Cn) with the end probability(end_probab) of all tokens upto the Span length (Cn+span). We use the (tf.matmul) operator for this step.
b. We use a hyperparameter - 'Span Length' to control the span for this computation. This span is the number of words between the Start and End token. We have used a Span of 20, since that is the average answer length based on the data analysis.
c. Since the previous step computes the joint probability between all the tokens in the context (C1..Cn), we must apply a condition that the Start token position must be before the End token. Any combination of tokens and their probabilities that do not satisfy this criteria are considered as invalid. We construct a similarity matrix ( context length X context length). This can be visualized as shown in Figure 3

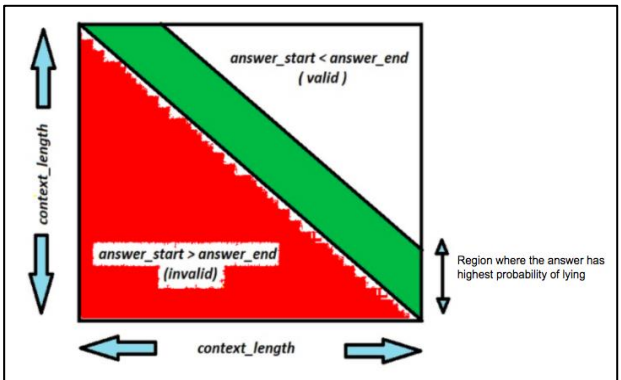

Fig.3. Similarity Matrix

Since Answer End Token > Answer End Token, the correct answer will always be in the matrix's upper part. Finally, we output the probability ( Y_probab) as a concatenation of the start and end positional token probabilities. From the above (Y_probab) to the actual value of Y_predict happens outside the model. We expect that to happen at the time of inference. We use the argmax function on our Y_probab, thereby getting a final array of Y_predict having the same dimension of Y_probab and being a sparse matrix with the predicted answers one-hot encoded.

d. *Custom Loss Function:* The custom loss function(logits_loss) computes the loss between the Predicted(Y_predict) and True values(Y_true). Since both these are encoded as a combination of Start and End tokens parameters, we compute the loss for the Start token and End tokens separately. After we obtain the loss for start token and end token values, we sum them to compute the Total loss[11].

e. *Optimization:* We have used 'Adamax' for gradient descent. It is a variant of Adam based on the infinity norm. We have selected Adamax as it is sometimes superior to Adam, especially in models with embeddings as per this Research

## III. RESULTS

Below table shows results of all RNN and Transformer models with their score

TABLE.1. MODEL PERFORMANCE

| | Model Name | F1(Ans) | EM(Ans) | F1(Plau Ans) | EM(Plau Ans) |
|---|---|---|---|---|---|
| 0 | LSTM Baseline | 29.00112098 | 28.22500192 | 0.2146227416 | 0.05371805694 |
| 1 | Deep LSTM + GloVe | 26.66643901 | 26.03023559 | 0.2763011224 | 0.06906607321 |
| 2 | Bi-LSTM + GloVe | 28.02196074 | 27.52666718 | 0.2811061296 | 0.1266211342 |
| ✅ 3 | Bi-LSTM + GloVe + C2Q Attention | 33.09511933 | 33.09416008 | 0 | 0 |
| 4 | Bi-LSTM + GloVe + Q2C-C2Q Attention | 29.62215708 | 29.13053488 | 0.2000388679 | 0.06906607321 |
| 5 | LSTM Baseline + Universal Sentence Encode | 24.54525456 | 22.95295833 | 0.6185752402 | 0.09976210575 |
| 6 | Bi-LSTM + Universal Sentence Encode | 32.02522636 | 31.54401044 | 0.06841133647 | 0.0230220244 |
| 7 | Bi-LSTM + C2Q Attention + Universal Sentence Encode | 30.69597542 | 30.12048193 | 0.1469459993 | 0.04220704474 |
| ✅ 8 | BERT + Cased_L-12_H-768_A-12 + DeepPavlov | 59.09692802 | 51.36213644 | 22.53928947 | 18.17972527 |
| 9 | BERT + Uncased_L-24_H-1024_A-24 + Huggingface | 57.51315339 | 49.76977976 | 22.1581681 | 17.75381782 |

For a simpler representation if we plot a bar graph and come to a conclusion that the BERT model using transfer learning from DeepPavlov gives the best results [7].

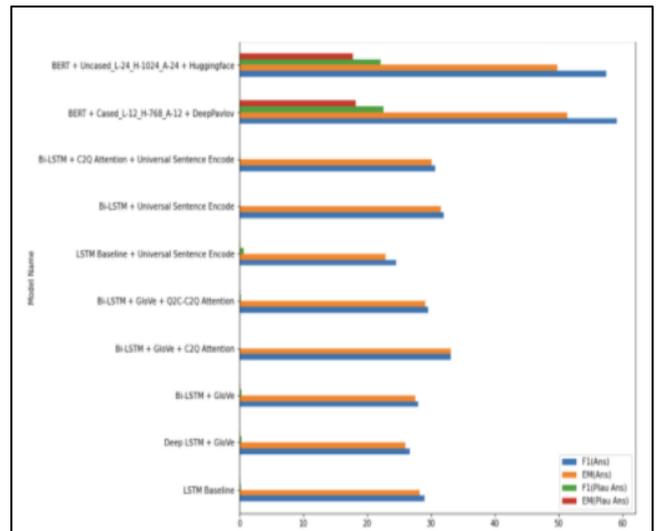

Fig.4. Model Performances

A baseline comparison of 2 metrics, one on Human Performance and another from the SQuAD 2.0 leader.

TABLE 2 MODEL PERFORMANCE COMPARED TO BENCHMARK

| | F1 | EM |
|---|---|---|
| Human Performance Stanford University (Rajpurkar & Jia et al. '18) | **89.452** | **86.831** |
| SA-Net on Albert (ensemble) QIANXIN | 93.011 | 90.724 |
| LSTM Baseline Model | 29 | 28 |
| biLSTM+GloVe+Context to Question Attention | 33.095 | 33.094 |
| BERT + Uncased_L-24_H-1024_A-24 + Huggingface | 57.513153 (+22.1581) | 49.769780 (+17.7538) |

The + 22.1581 and +17.7538 for the BERT model indicates match in plausible answers, which is the performance on unanswerable questions from the data set. This means that the BERT model is able to answer questions for which the data is not available.

## IV. DISCUSSION AND CONCLUSIONS

There are several reasons which we think the model performance could not match the benchmarks.

The model which beats human performance in SQuAD 2.0 leaderboard uses a Transformer based architecture with ensemble techniques. 2 of our models are based on RNN's. The LSTM Baseline model fails to capture language sequence information as a Transformer model does. LSTM baseline model also suffers from lack of bi-direction sequential information We have used "pre" as our padding to make data sequence length the same. However, as part of our embedding layer, we had to state mask=false - indicating that padded values should not be masked. This is done to make use of CuDNN LSTM. However, we think that model might be learning 0 as a value as well.

We attempted to add Attention as part of our BILSTM model. Generally, using the Attention mechanism should have given us a big boost, which we got by 4%. However, this Attention implementation is only from Context to Question. We attempted a bi-directional attention model from Question to Context and Context to Question. However, we saw a reduction in accuracy. We think that the implementation of bi-DAF is not on par with benchmarks. Other benchmark RNN models used a multi-embedding phase both at the phrase and character level on top of the word token level. We only could use the word level. Our epoch size on model training is 25. Other RNN benchmark models will have higher epochs. Hyper-parameter tuning needed more focus; we realized our learning rate parameter is not optimal. Our BERT based Transformer model, however, supersedes our RNN models by a considerable margin. However, fine-tuning the model with and maybe applying distillation in BERT could have helped.

Our models' primary limitation is that it targets only 'extractive answers', where the answer is always a continuous span of text in a paragraph. Our solution does not address 'abstractive answers' where the answer has to be inferred based on multiple information sources. Our solution pays 'attention' to simple reasoning skills like locating, matching or aligning information between query and context. Many real-life situations require the answer for a question to be extracted from multiple documents and then summarized.

To enhance our solution which should
- Enhance the training data size
- Increase the variety of Training data For example:
  - 'Cloze style questions' where the answer is a prediction of a missing word in a sequence;
  - 'Narrative QA', where high-level abstraction or reasoning is required to answer the questions
  - 'Multiple Choice Questions'
- Use other Transformer and SOTA models for language modelling tasks.